\documentclass[letterpaper, 10 pt, conference]{ieeeconf}  
\IEEEoverridecommandlockouts        
\usepackage{microtype}
\usepackage{times} 
\usepackage{amsmath} 
\usepackage{amssymb}  
\usepackage{multirow}

\usepackage[hidelinks,pdftex,pdfauthor={Justin Miller, Andres Hasfura, Shih-Yuan Liu, Jonathan P.\ How},pdftitle={Dynamic Arrival Rate Estimation for Campus Mobility on Demand Network Graphs}]{hyperref}
\usepackage[authormarkuptext=name,addedmarkup=bf,authormarkupposition=left]{changes}

\title{\LARGE \bf
Dynamic Arrival Rate Estimation for Campus Mobility on Demand Network Graphs}

\author{Justin Miller, Andres Hasfura, Shih-Yuan Liu, and Jonathan P.\ How
  \thanks{Laboratory of Information and Decision Systems,
    Massachusetts Institute of Technology, 77 Massachusetts Ave.,
    Cambridge, MA, USA {\tt\small \{justinm, hasfuraa, syliu, jhow\}@mit.edu}}%
}

\usepackage{mathtools}

\usepackage[labelfont=bf,font=small]{caption}
\usepackage{float}
\usepackage[labelfont=bf,font=small]{subcaption}

\usepackage{amssymb}
\usepackage[linesnumbered,ruled,vlined]{algorithm2e}

\usepackage{bbm}

\usepackage{balance}

\usepackage[capitalize]{cleveref}
\crefformat{equation}{(#2#1#3)}
\Crefformat{equation}{Equation~(#2#1#3)}
\Crefname{equation}{Equation}{Equations}

\begin{document}

\maketitle
\thispagestyle{empty}
\pagestyle{empty}

\begin{abstract}
Mobility On Demand (MOD) systems are revolutionizing transportation in urban settings by improving vehicle utilization and reducing parking congestion.
A key factor in the success of an MOD system is the ability to measure and respond to real-time customer arrival data.
Real time traffic arrival rate data is traditionally difficult to obtain due to the need to install fixed sensors throughout the MOD network.
This paper presents a framework for measuring pedestrian traffic arrival rates using sensors onboard the vehicles that make up the MOD fleet. 
A novel distributed fusion algorithm is presented which combines onboard LIDAR and camera sensor measurements to detect trajectories of pedestrians with a 90$\%$ detection hit rate with 1.5 false positives per minute.
A novel moving observer method is introduced to estimate pedestrian arrival rates from pedestrian trajectories collected from mobile sensors. 
The moving observer method is evaluated in both simulation and hardware and is shown to achieve arrival rate estimates comparable to those that would be obtained with multiple stationary sensors.
\end{abstract}

\section{Introduction}\label{sec:introduction}
Mobility On Demand (MOD) systems are revolutionizing transportation systems in urban settings by providing commuters access to transportation when needed without the need for private vehicle ownership.
Advantages of MOD systems include higher vehicle utilization rates and more sustainable urban land use through reduced parking spaces \cite{zhang_autonomous_2014}.
The main challenge for any MOD system is to provide the benefits of the private automobile experience while minimizing drawbacks that are introduced from using a shared resource \cite{mitchell_mobility_2008}. 
A major drawback for a customer relying on an MOD system is the potential for pickup delays that are not present in standard ownership models.
To mitigate this problem, an MOD system must leverage powerful and accurate demand prediction models to efficiently manage the spatio-temporal distribution of its fleet \cite{mitchell_mobility_2008}. 

Autonomous MOD systems composed of self-driving cars have been proposed as a means for controlling the coordination of the MOD fleet with respect to customer demand \cite{pavone_autonomous_2015}.
Previous work in this domain has focused on methods for autonomously repositioning vehicles within the MOD system in response to known customer arrival rates. 
The MOD system is modeled by either a fluid model \cite{pavone_robotic_2012} or a Jackson network model \cite{zhang_control_2016} in which customers arrive at stations on a network graph according to a Poisson process with constant arrival rate.
An optimal policy for system-wide vehicle coordination which meets customer demand is then developed.

\begin{figure}
	\center
	\includegraphics[width=1\columnwidth]{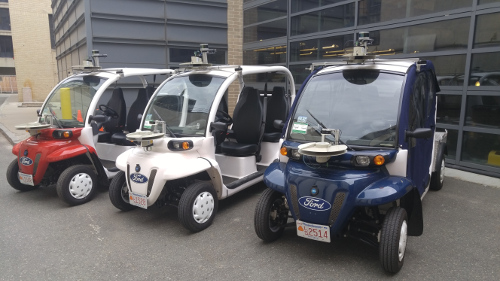}
	\caption{A fleet of MOD vehicles each mounted with camera and LIDAR sensors are used for pedestrian arrival rate estimation.}
	\label{fig:fleet}
\end{figure}

To build informative customer demand models, both historical databases for long-term prediction as well as real-time sensors for short-term fluctuations should be utilized \cite{mitchell_mobility_2008}.
While demand models can leverage previously served customers for historical input, obtaining real-time data is more challenging.
In autonomous MOD systems, each self-driving car is already equipped with a suite of real-time sensors that enable it to drive autonomously.
This work focuses on how sensors onboard MOD vehicles can be used to estimate real-time traffic data.
Specifically, MOD vehicles equipped with a suite of LIDAR and camera sensors typically found on self-driving cars, shown in \cref{fig:fleet}, are used to estimate the arrival rates of pedestrian traffic on a university campus.
In this setting, the traffic data can be used to inform MOD customer demand models, where customer ride request rates are expected to be correlated with overall traffic arrival rates.

The first challenge for using MOD vehicles as pedestrian traffic sensors is that of tracking pedestrians in real time, for which no single sensor solution exists.
Typically, a camera is used for pedestrian classification while LIDAR is used for tracking, and data from the two sensors must be fused together in a way that is robust to extrinsic calibrations.
Another challenge is that of estimating pedestrian arrival rates from vehicles that are moving alongside the pedestrian traffic.
Traditional traffic monitoring methods estimate arrival rates using data collected from stationary sensors located along roadways~\cite{_traffic_2013} by simply counting the number of passing pedestrians in a fixed time.
These fixed traffic sensors suffer from high infrastructure and maintenance costs, and full coverage within an MOD system is typically infeasible.
Moving sensors allow for greater coverage, but simple methods for estimating pedestrian arrival rates no longer apply because the measured pedestrian counts are dependent on the vehicle's location, orientation, and speed.
Previously explored moving observer methods are either unable to measure arrival rate information or are limited in the number of measurements they can make.

The contributions of this work are 
1) a novel distributed fusion scheme for pedestrian tracking using 2D LIDARs and RGB cameras that is robust to extrinsic calibration errors;
2) a novel moving observer method for estimating pedestrian arrival rates from mobile sensors; and
3) evaluation of arrival rate estimation accuracy of mobile sensors on both simulated and real data sets.

\section{Related Work}\label{sec:related_work}
\subsection{Pedestrian Detection and Tracking}
\label{sec:related_work_ped_tracking}
Pedestrian detection and tracking methods aim to track pedestrians in space, as opposed to just within an image frame. 
The two major sensor suites for pedestrian tracking are 3D LIDAR and camera plus 2D LIDAR. 
Discussion of single camera solutions is omitted due to typical inability to infer depth information.
Discussion of 2D LIDAR solutions without a camera is omitted because of work done by \cite{premebida_exploiting_2009-1} showing 2D LIDAR does not provide enough information to accurately classify pedestrians.

In \cite{kisku_multibiometrics_2011,cho_learned_2014}, a 3D LIDAR method is used to perform pedestrian detection and tracking, but because of speed limitations these methods provide too course of observations to perform tracking in cluttered environments.
In \cite{premebida_lidar_2007} a 2D LIDAR and camera are used where classification was performed individually on each sensor and a Bayesian fusion scheme produces a final classification. 
A similar approach is taken in \cite{premebida_lidar_2009,fotiadis_human_2013-1}, but with more modern alternatives for vision and LIDAR submodules. 
HOG+SVM is used as the vision classifier, hand made feature sets are proposed for LIDAR classification, and the authors experiment with different fusion techniques, such as centralized or decentralized. 
In \cite{premebida_fusing_2013}, the addition of semantic information is used to provide a prior for an object being a pedestrian. 
None of these papers exploit the temporal information available from multiple detections of the same pedestrian or multiple missing detections from non-pedestrians.

\subsection{Moving Observer Methods}
The current methods for automatic monitoring of pedestrian traffic \cite{_traffic_2013} rely on fixed sensors which can be cost prohibitive or infeasible for MOD system coverage.
Moving observer methods have been proposed for traffic monitoring using mobile sensors, although only for motorized traffic.
There are two main variations of moving observer methods: floating car methods and counting methods \cite{mulligan_uncertainty_2002}.
Floating car methods track a single vehicle as it moves with traffic to infer traffic stream characteristics such as the speed of vehicles along roadways.
In \cite{yim_investigation_2001} and \cite{cathey_transit_2001}, probe vehicles equipped with GPS or cellular tracking devices are used to estimate traffic speeds and travel times for vehicles traveling along various road segments.
In \cite{chen_gaussian_2015}, MOD vehicles are used as active sensing probes for estimating traffic speed data.
The floating car method has the fundamental limitation that primarily only speed data can be estimated since only the probe vehicle itself is actually observed; traffic arrival rate information is not available.
The method is also not applicable to vehicles operating in pedestrian traffic networks because the probe vehicles and pedestrians are typically moving at different speeds.

Counting methods consist of a moving observer counting the number of vehicles that are overtaking or are being overtaken by the moving observer.
These methods estimate the traffic flow rate along road networks, which is equivalent to measuring traffic arrival rates.
In \cite{wardrop_method_1954}, a vehicle is driven along a road segment twice, once in each direction, and an observer in the vehicle records the number of oncoming vehicles, the number of overtaken vehicles, the number of overtaking vehicles, and the average speed of the vehicle.
The traffic flow rates in both directions are then computed from the recorded data.
A one-way moving observer method is presented in  \cite{mulligan_uncertainty_2002} which only requires a single pass of a road segment, but it only estimates one-way flow rates in the opposite direction to the observer.
Counting methods are limited in the amount of data that can be collected because vehicles must traverse an entire road segment in both directions to obtain flow estimates for both directions of traffic.
The idea of using self-driving vehicle sensors to estimate flow rates is presented in \cite{redmill_using_2011}, but the focus of the work was on developing the detection capabilities of the vehicle and flow rate estimation methods were not addressed.

Compared to previous work, the moving observer method proposed in this paper allows arrival rate estimation for non-motorized traffic on any portion of the roads within the vehicle's sensing region, whether the vehicle is traveling with, against, or adjacent to the traffic.

\section{Data Fusion for Pedestrian Detection and Tracking}\label{sec:ped_tracking}

This section details the framework for fusing LIDAR and camera sensor data into pedestrian trajectory data using sensors mounted on vehicles driving on a university campus.

\subsection{Hardware}
Three Polaris GEM vehicles are each equipped with three Logitech C920 cameras and two SICK LMS151 LIDARs as shown in \cref{fig:fleet}.
The cameras are operating at 30 fps with a resolution of 640x360 and are mounted to cover 160 degrees around the front and sides of the vehicle. 
The LIDARs are operating at 50Hz with a 50m range, 270$\deg$ field of view, and 0.5$\deg$ angular resolution.
Both LIDARs are facing the forward direction of the vehicle, one is mounted on the hood and the other on the roof.
The vehicles are also equipped with Velodyne VL-P16 3D LIDAR sensors, but these are currently not used to preform pedestrian tracking for computation reasons as stated in \cref{sec:related_work_ped_tracking}.

\subsection{Localization}
Collecting accurate trajectories of pedestrians requires the MOD vehicle to be localized on a map.
All localization is performed using laser scans from the roof-mounted SICK LIDAR.
The Robot Operating System (ROS) \cite{quigley_ros_2009} is used for managing sensor data and perception algorithms.
Odometry for the vehicle is estimated using the Hector SLAM ROS package which performs laser scan registration on multi-resolution grid maps \cite{kohlbrecher_flexible_2011}.
An occupancy grid map of the environment is obtained using the Gmapping ROS package which uses a particle filter based SLAM approach \cite{grisetti_improved_2007}.
Real-time localization of the vehicle is achieved using the AMCL ROS package which uses a particle filter to track the pose of the vehicle with respect to the generated map \cite{thrun_robust_2001}.

\subsection{Pedestrian Tracking}
\begin{figure}
  \center
  \includegraphics[width=1\columnwidth]{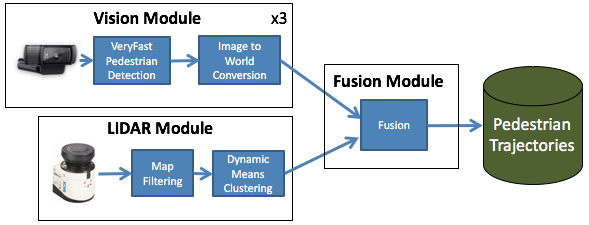}
  \caption{Schematic of overall system architecture illustrating how the vision, LIDAR, and fusion modules are combined to generate pedestrian trajectories.}
  \label{fig:overall_system}
\end{figure}

There are three modules used for pedestrian tracking: the LIDAR module provides trajectories of objects in the map frame, the vision module detects pedestrians in the image frame, and the fusion module classifies tracked objects as pedestrians by fusing the outputs of the LIDAR and vision modules.
An overview of the system architecture is shown in \cref{fig:overall_system}.

\subsubsection{LIDAR Module}
The laser scans of the hood-mounted LIDAR are first transformed into the map frame using the vehicle localization.
Points in the scan that overlap with the statically occupied cells in the pre-built occupancy grid map are filtered out.
Sequential measurements from the filtered laser scans are then clustered together using Dynamic Means \cite{campbell_dynamic_2013-1}, which is a fast, general purpose clustering algorithm designed to captured the temporal and spatial evolution of clusters. 
The sequence of Dynamic Means position estimates for each cluster is used to directly provide cluster trajectory data in the map frame with no additional processing.

\subsubsection{Vision Module}
Image data from the three cameras on the vehicle are all streamed through a VeryFast pedestrian detector \cite{benenson_pedestrian_2012-1}, which produces bounding boxes indicating the locations of pedestrians in the images. 
The left, middle, and right edges of the pedestrian's bounding box are converted into a set of bounding box vectors that are projected into the map frame using extrinsic calibration data.
The SVM based VeryFast detector is able to achieve high frame rate detections on the order of 90 fps, which makes it well suited for real time applications.

\subsubsection{Fusion Module}
\begin{figure}
  \center
  \includegraphics[width=1\columnwidth]{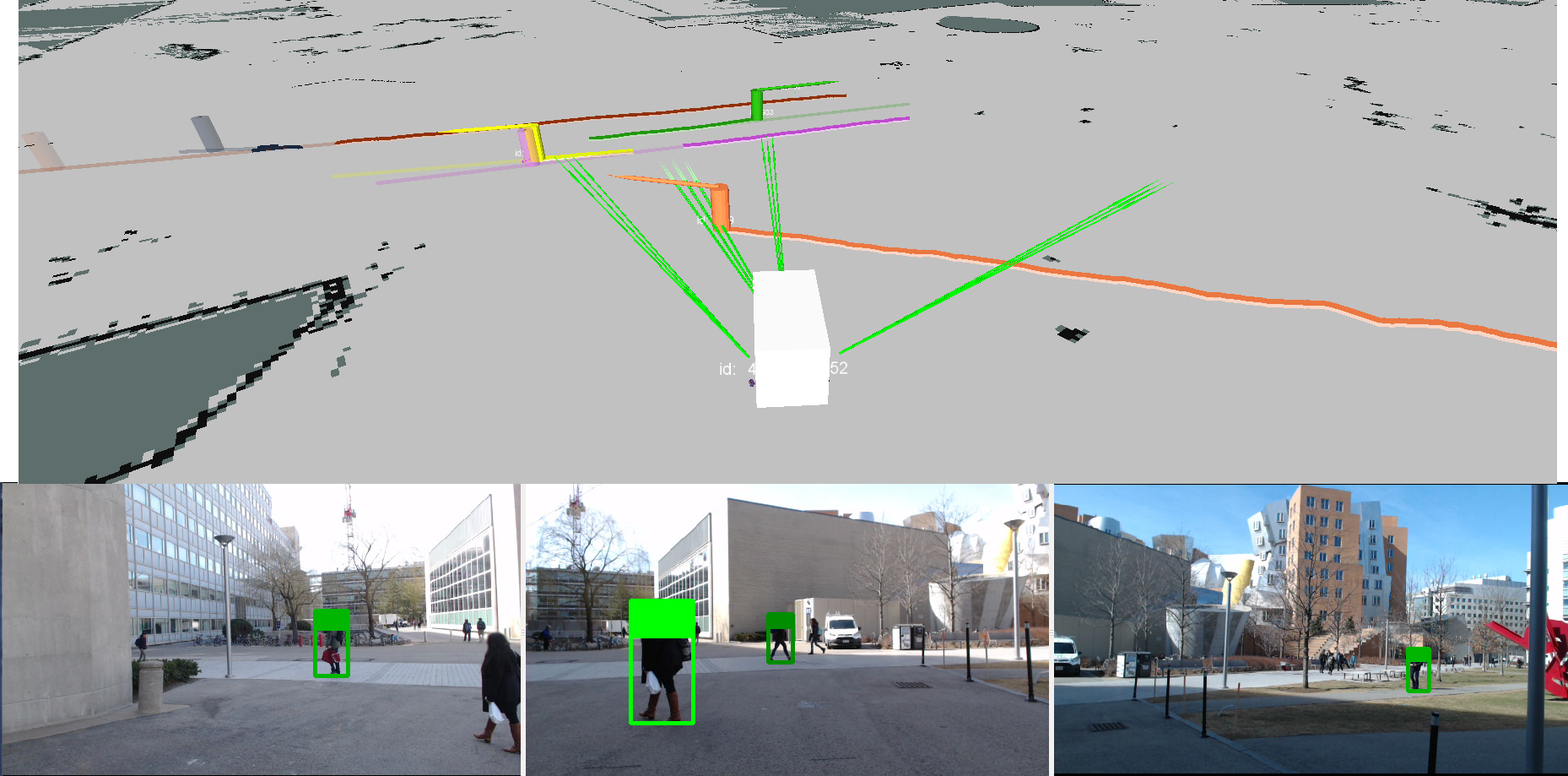}
  \caption{The pedestrian tracking framework is implemented in real time on an MOD vehicle. (Top) The white box represents the vehicle, green arrows represent the vectors of pedestrian detections projected into the map frame, cylinders represent detected pedestrians with paths trailing. (Bottom) The images from the three cameras with bounding boxes of the detected pedestrians outlined in green (faces covered for privacy reasons).}
  \label{fig:system_visualization}
\end{figure}

Bounding box vectors from the vision module are aligned with cluster positions from the LIDAR frame through extrinsic calibration data. 
In a typical maximum likelihood fusion (MLF) method, the cluster most aligned with a bounding box vector receives a ``hit" count, while others receive nothing. 
However, minor error in intrinsic and extrinsic calibration can cause a significant drop in detection performance. 
Instead, a novel distributed fusion (DF) method is proposed where partial hits are assigned to multiple clusters within range of a bounding box vector.
A cluster's partial hit count $h$ is provided by
\begin{equation}
  h = e^{ -\frac{d^2}{2\sigma} } \label{eq:prob_fusion} \, ,
\end{equation}
where $d$ is the sum of the angular distances between the set of bounding box vectors and a cluster, and $\sigma$ is a tunable variance parameter which controls the magnitude of the hit with respect to alignment distance.
In DF, clusters which are not perfectly aligned with a bounding box can still receive a relatively high partial hit count, which makes the method more robust to extrinsic calibration errors.
Over time, clusters continually close to bounding box vectors will receive increasingly larger hit counts, while those farther away will also receive decreasingly smaller hit counts. 
Because of this, non-pedestrian clusters should become less likely over time to have a high hit count, resulting in fewer false detections.
In both MLF and DF methods, clusters that have accumulated high enough hit counts are labeled as pedestrians and the cluster trajectories are recorded.
\cref{fig:system_visualization} shows the pedestrian tracking framework on an MOD vehicle. 
Pedestrians bounding boxes shown in the images are projected out into the map frame where they are fused with clusters.

\section{Pedestrian Arrival Rate Estimation}\label{sec:ped_traffic}

The pedestrian trajectories that are observed from moving MOD vehicles can be used to estimate pedestrian arrival rates within an MOD network graph.

\subsection{Pedestrian Network Graph}
A pedestrian traffic network is modeled as a directed graph denoted by $\mathcal{G} = (\mathcal{N},\mathcal{L})$, where $\mathcal{N} = \{ n_1, \dots, n_{N} \}$ is a set of $N$ nodes, and $\mathcal{L} = \{l_1, \dots, l_{L}\}$ is a set of $L$ directed link edges each taking the form of an ordered pair of nodes $l = (n_i, n_j) \in \mathcal{N}^2$.
A node $n \in \mathcal{N}$ represents a region in the network graph where pedestrians can arrive, leave, or change directions.
A link $l \in \mathcal{L}$ represents a path that pedestrians are constrained to walk along when traveling between nodes, such that a pedestrian walking along link $l = (n_i, n_j)$ must travel directly from $n_i$ to $n_j$.
For each pair of connected nodes, there are exactly two directed links connecting them, pointing in opposite directions.

\begin{figure}[t]
	\vskip 0.1in
	\centering
	\includegraphics [width=0.45\textwidth]{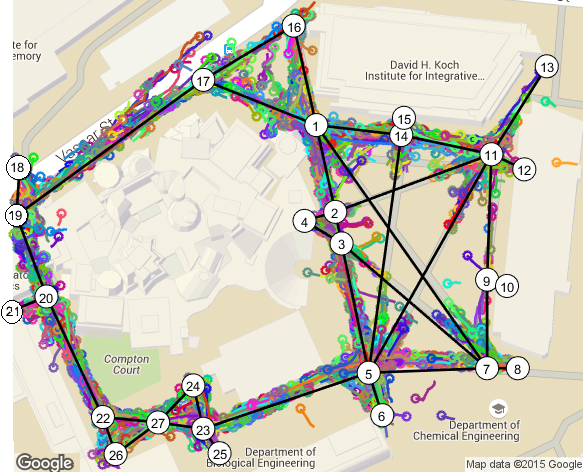}
	\caption{The pedestrian traffic network on MIT campus is determined from pedestrian trajectories overlaid on a map. The network graph is composed of 27 nodes and 74 directed links.}
	\label{fig::network_map}
	\vskip -0.2in
\end{figure}

The structure of the network graph can be manually estimated from the pedestrian trajectory data collected from MOD vehicles.
Over 16000 pedestrian trajectories were collected from a fleet of MOD vehicles that had driven over a total of 30 vehicle hours.
The trajectories are overlaid on a map of the MOD region and nodes and links are manually assigned to best fit the data.
Typically, links are identified from parallel trajectories and pathways on the map.
Nodes are identified from intersecting trajectories and known origins and destinations on the map.
\cref{fig::network_map} shows the traffic network graph generated from pedestrian trajectories overlaid on a map.

A network route $r \in \mathcal{F}$ is defined as a set of directed links $\mathcal{L}_{r} \subseteq \mathcal{L}$ which connect an origin node $o \in \mathcal{O}$ to a destination node $d \in \mathcal{D}$, where $\mathcal{O} \subseteq \mathcal{N}$ is the set of nodes where pedestrians can enter the network, $\mathcal{D} \subseteq \mathcal{N}$ is the set of nodes where pedestrians can exit the network, and the set $\mathcal{F}$ contains all minimum length routes between all possible origin-destination pairs on the MOD graph.

Pedestrians are modeled as arriving in the system with a predetermined route $r$ according to a Poisson process; the arrival rate  parameter for route $r$ is denoted by $\lambda_{r} \in \mathbb{R^+}$.
Pedestrians traveling along the routes will then generate arrivals at the origin nodes for links along the route.
Following the rule of superposition for Poisson processes, the arrivals at links $l$ will also be distributed according to a Poisson process with arrival rate parameter $\lambda_l \in \mathbb{R^+}$ equal to
\begin{equation}
\lambda_{l} = \sum\limits_r\lambda_{r} \cdot I_{\mathcal{L}_r}(l),
\end{equation}
where 
\begin{equation*}
I_{\mathcal{L}_r}(l) :=
\begin{cases}
1 &\text{if } l \in L_r, \\
0 &\text{if } l \notin L_r.
\end{cases}
\end{equation*}

In general, it is not possible to continuously observe pedestrians traveling along an entire route since an observer would have to travel the same route at the same time for every pedestrian.
Instead observations are made only at the link level where an observed pedestrian is assumed to travel the whole length of the link.
For this reason, this work aims to only estimate the link level arrival rates for each link; that is the set $\mathbf{\Lambda} = (\lambda_1, \dots , \lambda_L)$. 
Estimating the route arrival rates from the set of link arrival rates is a separate active research field known as Origin-Destination Matrix Estimation, and the interested reader is referred to \cite{abrahamsson_estimation_1998}.

\subsection{Arrival Rate Estimation from a Moving Observer}
Arrival rates on links can be estimated from pedestrian trajectory data collected by moving observers through the use of pedestrian velocities and a known sensing field of view. 
Pedestrian velocities are estimated by differentiating pedestrian trajectory positions with respect to time.
The size and location of the sensing field of view are determined through sensor testing and vehicle localization, respectively.
It should be noted that the following analysis is worded for pedestrian sensing but can be applied to non-pedestrian traffic as well, given adequate sensing capabilities.

\subsubsection{Observation Method}
Consider a link in the network for which pedestrians arrive at its origin node according to a Poisson process with link arrival rate $\lambda \in \mathbb{R^+}$ and then travel along that link.
At time $t_p \in \mathbb{R^+}$, a pedestrian $p$ arrives at the link's origin node and travels along the link with constant velocity $v_p \in \mathbb{R^+}$.
At time $t \ge t_p$, the location $x_p \in \mathbb{R^+}$ of the pedestrian relative to the link's origin node will be $x_p = v_p (t-t_p)$.

An MOD vehicle in proximity of the link at $t$ will observe any pedestrians that fall within its sensing field of view, along with their velocity $v_p$.
The portion of the link within the sensing field of view has length $d_{obs} \in \mathbb{R^+}$ and is bounded by locations $\{x_1,x_2\}$, where $x_1 = x_2 + d_{obs}$.
For a pedestrian to be within the sensing region at time $t$, it must hold that $x_p \in [x_2, x_1]$.
Using the pedestrian's velocity, the bound on the pedestrian's arrival time at the link's origin node $t_p \in [t_1, t_2]$ can be determined where $t_1 = t- \frac{x_1}{v_p}$ and $t_2 = t- \frac{x_2}{v_p}$.
By extension, the vehicle would observe $n$ pedestrians at time $t$ if each pedestrian left the link's origin node between times $t_1$ and $t_2$ traveling at the same speed $v_p$.

Knowledge of the sensing region bounds $\{x_1,x_2\}$ and pedestrian velocity information $v_p$ can be used to estimate a window of \emph{projected observation times} $\{t_1,t_2\}$ within which $n$ pedestrians are estimated to have arrived in the network.
The count of $n$ pedestrians and the estimated arrival time period $\tau  \in \mathbb{R^+}$ defined as $\tau = t_2-t_1 = \frac{d_{obs}}{v_p}$ can then be used to estimate the Poisson arrival rate.

In practice, not all pedestrians will travel at the same speed in the network.
In the case where $n$ pedestrians are observed in the sensing region with different velocities $\{v_1 \dots v_n\}$, an approximation of $v_p$ is made using an average speed. 
In this case, it is more appropriate to use the space mean speed, as done in \cite{mulligan_uncertainty_2002}.
Unlike the arithmetic mean speed which averages over time, the space mean speed averages over distance and is defined as 
\begin{equation}
	\bar{v}_p = \displaystyle \frac{n}{ \sum_{i=1}^{n} v_i^{-1}}\, ,
\end{equation}
In the $n=0$ case, an expected pedestrian speed $\tilde{v}_p$ is used based on previously observed pedestrian velocities.

\subsubsection{Poisson Arrival Rate Estimation}
The Poisson rate parameter $\lambda_l$ can be estimated using the set of $N_o$ independent observations consisting of counts $[n_1 \dots n_{N_o}]$ and estimated arrival time periods $[\tau_1 \dots \tau_{N_o}]$. 
As in \cite{engelhardt_events_1994}, the maximum likelihood estimator (MLE) for a Poisson process is used, which is an efficient, unbiased estimator for $\lambda$.
The Poisson MLE is known to be 
\begin{equation} 
	\hat{\lambda}_l = \frac{N_c}{T_c} \, , 	\label{eq:mle}
\end{equation}
where $N_c = \sum_{i=1}^{N_o} n_i$ is the total number of counts and $T_c = \sum_{i=1}^{N_o} \tau_i$ is the total arrival time period.
The lower and upper confidence interval bounds for the estimate, $\lambda_L$ and $\lambda_U$, which indicate that the true value is within the bounds with $100(1-\alpha)$ confidence, are 
\begin{align} 
\lambda_L &= \frac{ \chi^2_{\alpha/2}(2N_c)}{2T_c}\, , \label{eq:lb} \\ 
\lambda_U &= \frac{ \chi^2_{1-\alpha/2}(2N_c+2)}{2T_c}\, , \label{eq:ub}
\end{align}
where $\chi^2_{p}(\nu)$ is the $p$-th quantile of the $\chi^2$ distribution with $\nu$ degrees of freedom.

\subsubsection{Independent Observations}
A criteria imposed on the data for estimating $\hat{\lambda}_l$ is that each observation must be independent, which requires that the projected observation times do not overlap.
As the vehicle moves, the projected observation times may overlap based on the relative speeds of the vehicle, pedestrians, and sampling rate of the data.
For example, in the limiting case where the vehicle is moving with the traffic flow and at the same speed, the projected observation times of the data will not change.
If the vehicle is making observations at a fixed sampling rate, it will generate multiple observations all with overlapping intervals.
To eliminate sampling rate factors from affecting the estimation results and ensure that the data is independent, new measurements with projected observation times overlapping with previous times are discarded.

\section{Experiments and Results}\label{sec:results}

\subsection{Pedestrian Detection and Tracking}
\begin{figure}
  \centering
  \includegraphics[width=0.5\textwidth]{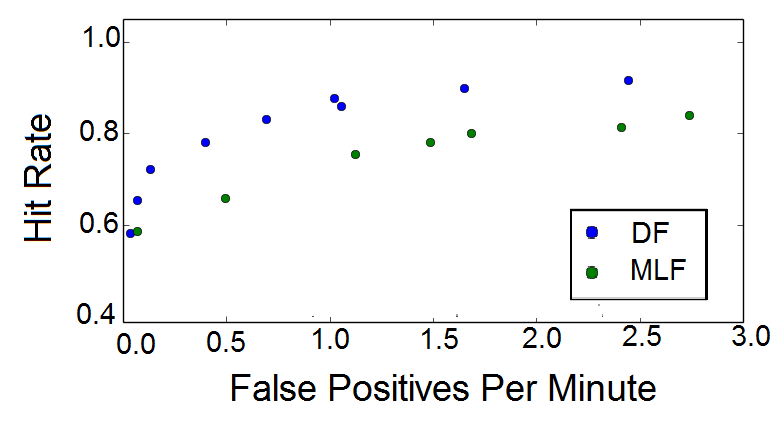}
  \caption{The ROC curve compares the performance of MLF and DF fusion methods on the recorded dataset.}
  \label{fig:old_vs_new_all}
\end{figure}

The receiver operating characteristic (ROC) performance of MLF and DF pedestrian tracking methods are compared on a ground truth dataset, as shown in \cref{fig:old_vs_new_all}.
The ROC curve is generated by varying the cluster hit threshold for which clusters are classified as pedestrians.
The x axis of the ROC curve shows the number of false positives per minute and the y axis shows the percent of pedestrians who's trajectories are correctly identified from the data set.
The dataset was recorded over a period of 22 minutes and 237 pedestrians were manually identified and labeled. 

DF outperforms MLF at all points on the ROC curve.
At the chosen ROC operating point corresponding to 1.5 false detections per minute, DF is able to identify 214 of 237 pedestrians correctly compared to the 190 of 237 for MLF, both with 33 false detections. 
The 90$\%$ hit rate for DF significantly outperforms the 80$\%$ hit rate of the MLF approach due to the improved robustness against false detections and extrinsic calibration errors.

\subsection{Arrival Rate Estimation}
In this section, the moving observer method is used to estimate pedestrian link arrival rates using vehicles traversing the network graph.
The accuracy and confidence of the arrival rate estimates are evaluated both with hardware and in simulation.

\subsubsection{Hardware Experiment}
A hardware experiment was performed using a manually driven MOD vehicle equipped with camera and LIDAR which provide a sensing region of 20 m range and 160 $\deg$ field of view.
To establish ground truth, an equally equipped MOD vehicle was parked along the link between nodes 20 and 22 and served as a stationary link counter.
While the moving vehicle drives, the DF method is used to obtain pedestrian trajectories and the moving observer method is used to estimate the pedestrian arrival rates in the network.
A demonstration video is available at \url{https://www.youtube.com/watch?v=skJQEiG-Hxg}.
The vehicle was driven at speeds between 2.5 and 4.5m/s for a time period of 1.5 hours, during which time the pose of the vehicle was always known through localization.
In this experiment, the vehicle was driven such that it attempted to cover all links equally.
Note, the process of having vehicles serve customers while also attempting to fully cover all links in the network, and the exploration vs. exploitation trade-offs that are incurred, is a separate interesting problem left for future work.

\begin{figure*}[t]
	\centering
	\hspace*{\fill}%
	\begin{subfigure}{0.4\textwidth}
		\centering
		\includegraphics[width=1\columnwidth]{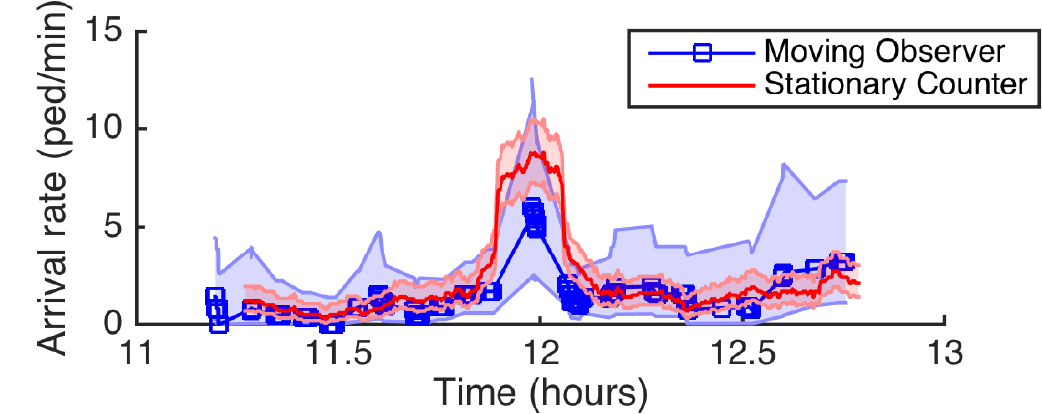}
		\caption{Arrival rate estimate comparison for pedestrians traveling from node 20 to node 22.}
		\label{fig::arrival_rate_57}
		\includegraphics[width=1\columnwidth]{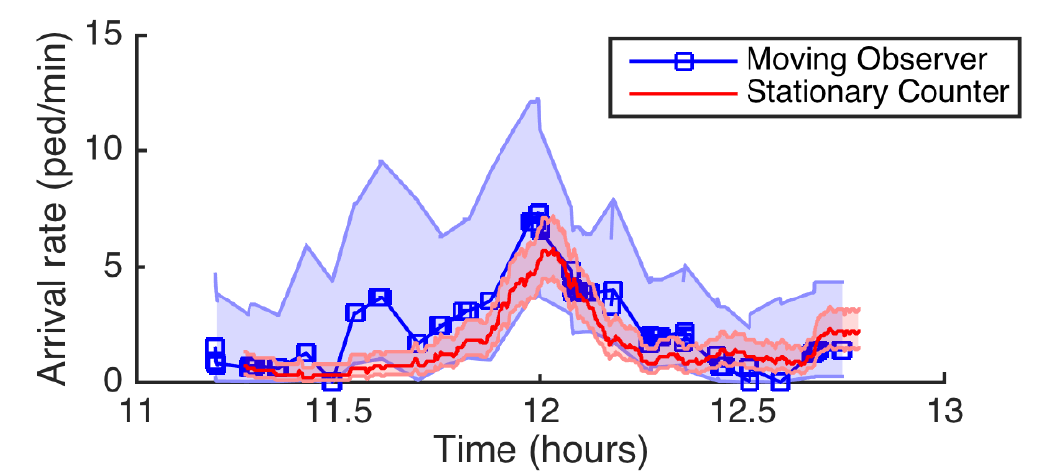}
		\caption{Arrival rate estimate comparison for pedestrians traveling from node 22 to node 20.}
		\label{fig::arrival_rate_59}
	\end{subfigure}
	\hfill%
	\begin{subfigure}{0.45\textwidth}
		\includegraphics[width=1\columnwidth]{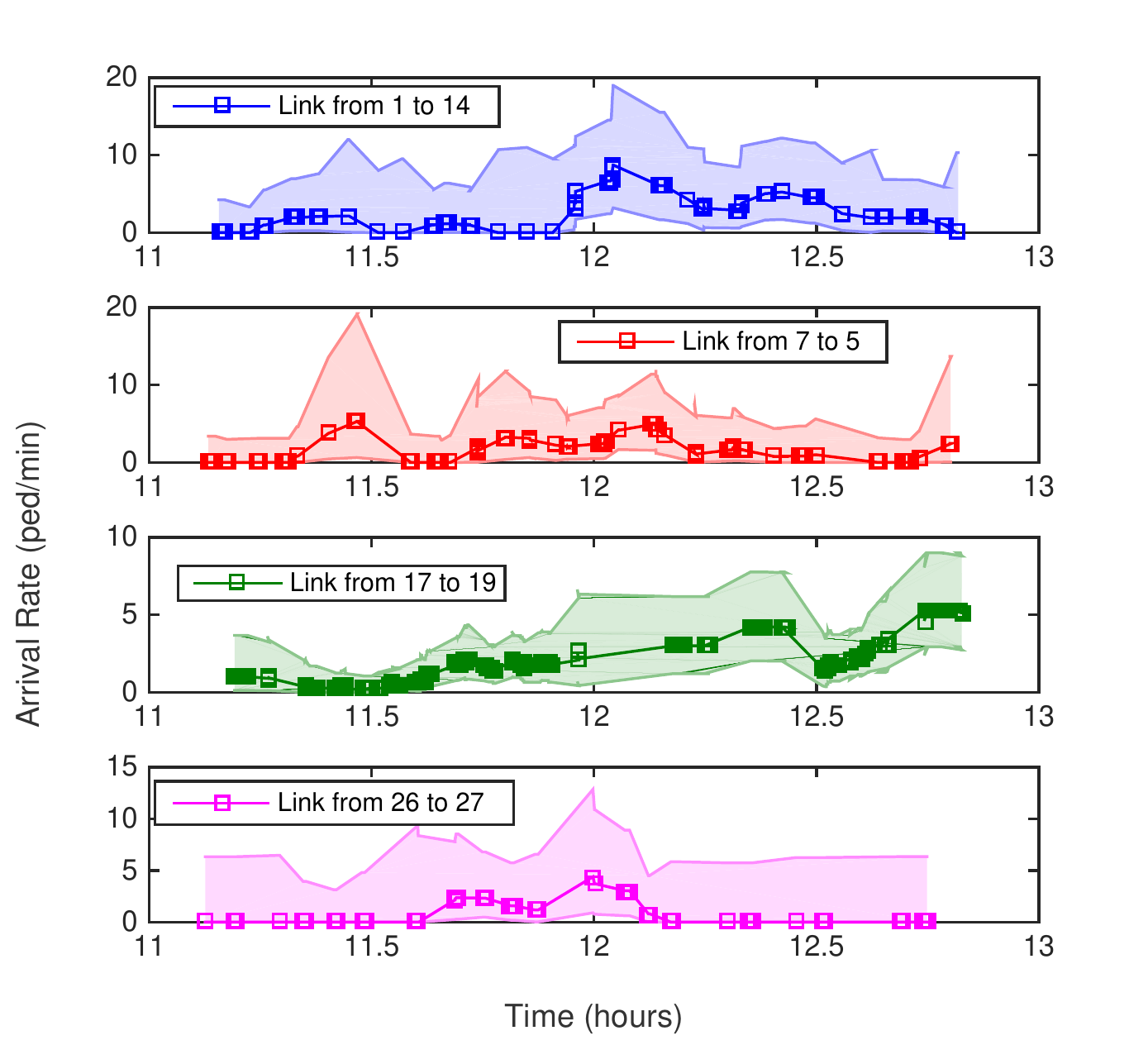}
		\caption{Arrival rate estimates for pedestrians traveling along four additional links in the network.}
		\label{fig::four_links}
	\end{subfigure}
	\hspace*{\fill}%
	\caption{Arrival rate estimates with a moving average filter applied, where arrival rates are estimated from data within a 10 minute interval centered at each point in time. The shaded regions indicates the 90\% confidence interval.}
\end{figure*}

\cref{fig::arrival_rate_57} shows the estimated arrival rates from both the moving observer and the stationary counter for pedestrians traveling from node 20 to 22. 
Similarly, \cref{fig::arrival_rate_59} shows the estimated arrival rates in the opposite direction, from node 22 to 20.
The time profile of the arrival rates is generated using a moving average filter, where arrival rates and confidence intervals are computed from data within a 10 minute interval using \cref{eq:mle,eq:lb,eq:ub}.
The time profiles indicates that the arrival rates generally vary over time, with a large increase around 12~PM, which is known to be common for these two links due to the change of classes that occurs then.
The moving observer only makes sparse measurements of the link, but the estimated arrival rate means at each measurement are generally consistent with those from the stationary counter.
The confidence intervals for the moving observer method estimates are much larger, which is expected due to the smaller amount of effective viewing time in the 10 minute window.
During that 10 minute window, the moving observer traversed other areas of the network graph.
Unlike the stationary counter, the moving observer was able to estimate similar arrival rate profiles for every other link in the network, four of which are shown in \cref{fig::four_links}.
In an MOD setting, less confident estimates of the relative arrival rates among all links is arguably more useful than very confident estimates on just a few links because it provides insights into which locations in the network graph have the most arrivals at any given time.
More confident estimates are achievable with the moving observers, however, if multiple MOD vehicles are traversing the network graph at the same time and their arrival rate observations are combined.
Similarly, through management of an MOD fleet, more confident estimates can be obtained by having vehicles spend more time on certain links than others.
In the limiting case where a vehicle is parked, the estimates would be equivalent to a stationary link counter.

\subsubsection{Simulation Experiments}
A simulator was developed to test the moving observer arrival rate estimates against multiple stationary counters on all links in the network.
In the simulator, a vehicle and pedestrians travel through the same traffic network that was shown in \cref{fig::network_map}.
Pedestrians arrive at the origin nodes of links according to a Poisson process with constant arrival rate and travel the link at constant velocity.
The vehicle travels at constant velocity and is driven such that when it reaches the end of a link, it transitions to whichever connecting link that has been visited the least previously.
This driving strategy is designed to facilitate balanced traversal of all links.
Pedestrians within the sensing region of the vehicle will be tracked and their trajectories recorded. 
After a fixed amount of time, the arrival rate estimates and corresponding confidence intervals from the moving observer are then computed using \cref{eq:mle,eq:lb,eq:ub}. 

\begin{figure}[t]
	\centering
	\includegraphics [width=0.8\columnwidth,trim=3 0 10 0,clip]{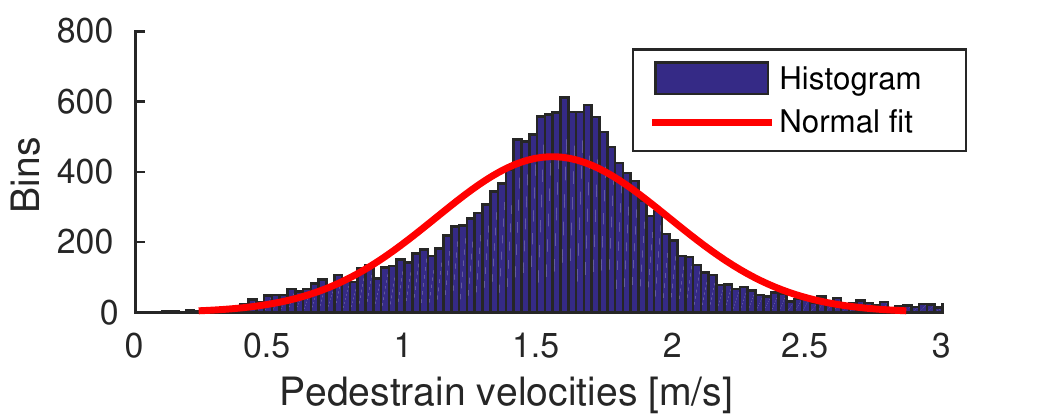}
	\caption{Histogram (blue) shows the measured velocities from 16632 pedestrians. The Normal fit (red) is used to generate random samples of pedestrian velocities in simulation.}
	\label{fig::velocity_bins}
\end{figure}

\begin{figure}[t]
	\centering
	\includegraphics [width=0.8\columnwidth,trim=5 0 10 0,clip]{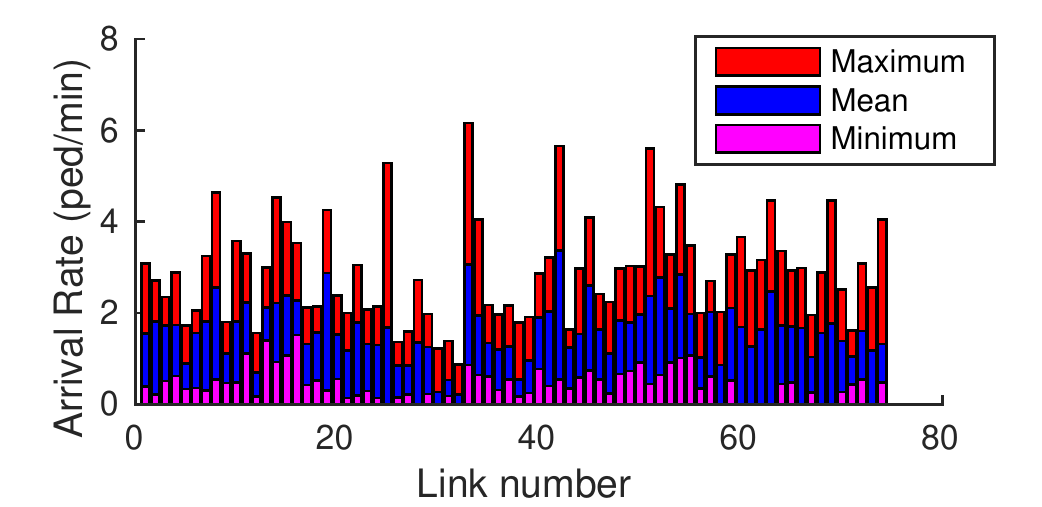}
	\caption{Arrival rates on each link in the network were estimated from an MOD vehicle using the moving observer method. Data was collected during the same 2 hour period between 11~AM and 1~PM and averaged over the course of 20 days.}
	\label{fig::average_real_flow_rates}
\end{figure}

The simulation parameters for the pedestrian velocities, the true pedestrian arrival rates, the vehicle velocity, and the sensing region of the vehicle are chosen to match collected real world data.
Pedestrian data was collected from an MOD vehicle operating on the real world network graph over the course of 20 days between the hours of 11~AM and 1~PM.
A histogram of pedestrian velocities from over 16000 pedestrian trajectories is shown in \cref{fig::velocity_bins}. 
Based on this, simulated pedestrian speeds are sampled from a normal distribution with mean speed of 1.5 m/s and standard deviation of 0.4 m/s.
The expected arrival rates in the network are determined by applying the moving observer method to data collected from the MOD vehicle. 
\cref{fig::average_real_flow_rates} shows the 20 day averaged arrival rates for each link, from which a nominal arrival rate of 1.62 ped/min was chosen for use in simulation.
The vehicle velocity is set to be a constant 3.5m/s, which is consistent with the speeds which are driven on the university campus.
The sensing region of the vehicle is chosen to be a 20m radius with 160$\deg$ field of view, as was determined from testing of the MOD sensors.

\begin{figure}[t]
	\centering
	\includegraphics [width=0.5\textwidth]{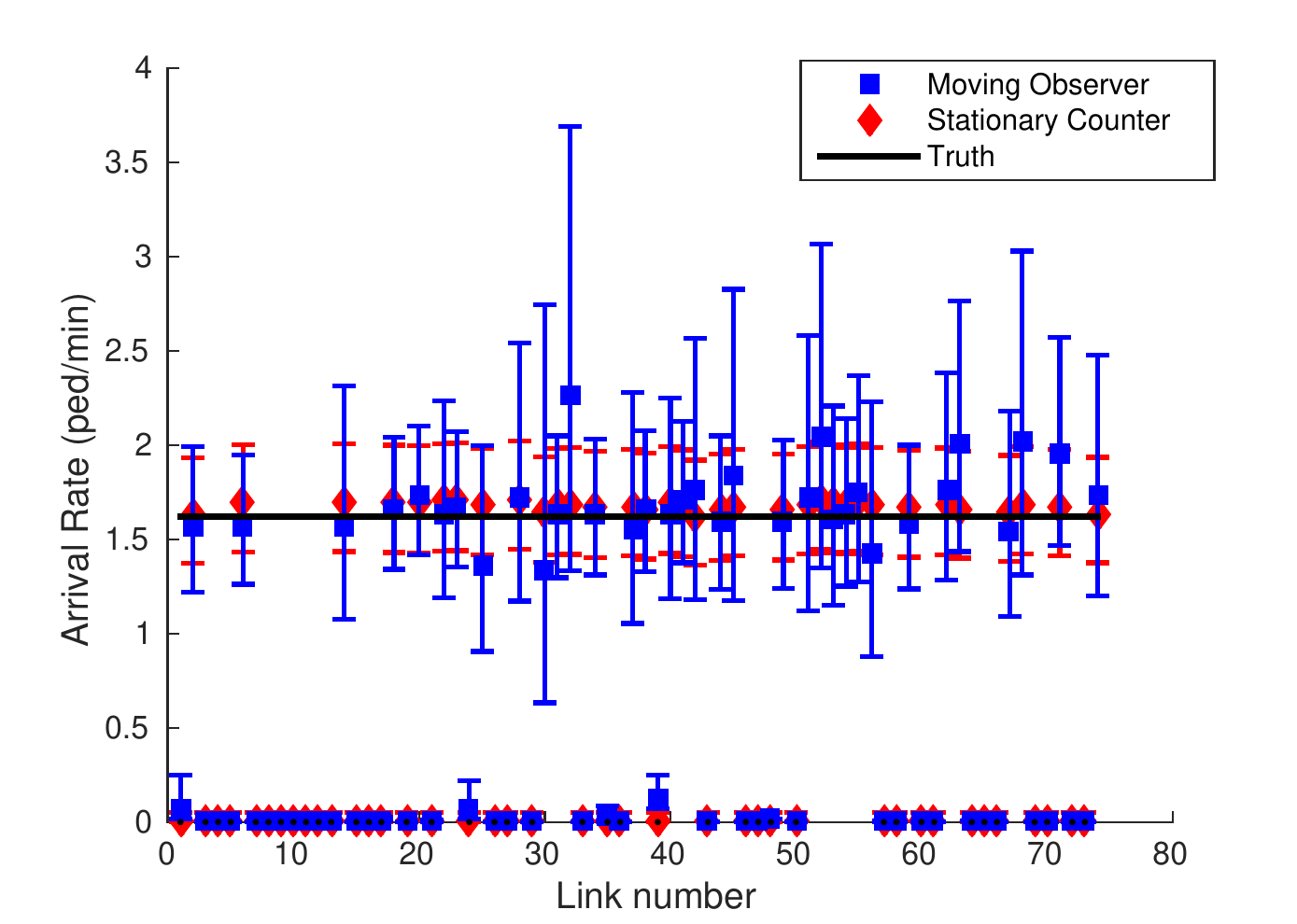}
	\caption{The results from a full network simulation compare the moving observer method (blue) with a stationary counter (red) and ground truth (black). The results are averaged over 100 simulated runs and 90\% confidence intervals are shown. }
	\label{fig::34_routes_3600}
\end{figure}

The simulator was run on the full pedestrian network with 34 active pedestrian links each having the same constant arrival rate.
In order to ensure that the vehicle can make a significant number of observations of each link, 1 hour of driving was simulated. 
\cref{fig::34_routes_3600} shows the results comparing the moving observer method with both a stationary counter and ground truth.
The results show that a single moving observer is able to generate reasonable arrival rate estimates for most links in the network. 
The estimates and confidence interval from the stationary counter indicate the best confidence bound that could be achieved by the moving observer method, which would occur if it was parked on the link.
To achieve this, however, there would need to be a stationary counter on every pair of directed links in the network, in this case 37 stationary counters were used.
In contrast, the moving observer method is able to generate an estimate for all of the links using only a single vehicle.
This, however, comes at the cost of less certainty in some estimates as indicated by the larger confidence intervals.

Those links with highest uncertainty correspond to the shortest links in the networks.
While the vehicle attempts to make an equal number of trips to each link, the effective observation time will be dependent on the length of the link and the vehicle's velocity.
This indicates that methods for actively sensing links should consider spending more time on shorter links by either visiting them more often or by reducing the vehicle's velocity while traversing them.

\begin{figure}[t]
	\centering
	\begin{subfigure}{0.5\textwidth}
		\centering
		\includegraphics[width=1\columnwidth]{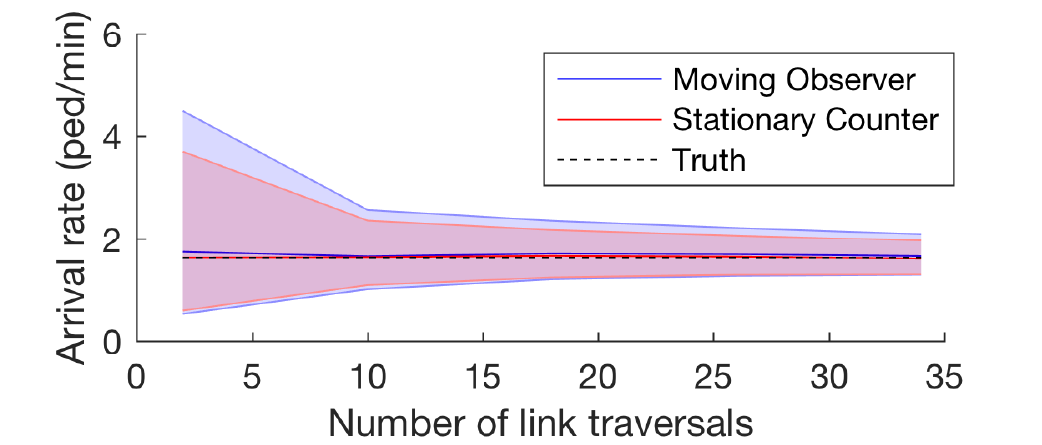}
		\caption{Arrival rate estimates for varied number of traversals of the link.}
		\label{fig::varying_num_visits}
	\end{subfigure}\\
	\begin{subfigure}{0.5\textwidth}
		\centering
		\includegraphics[width=1\columnwidth]{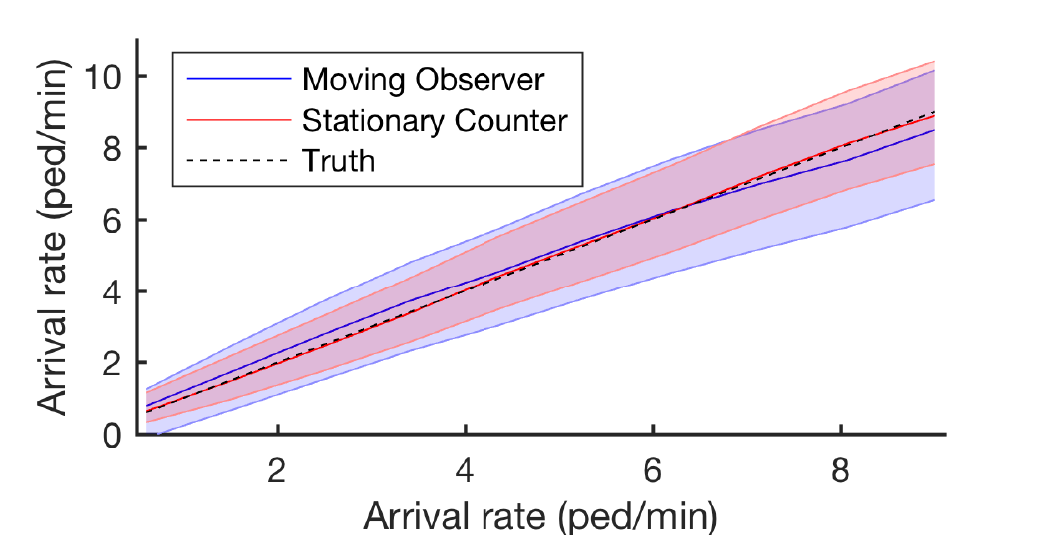}
		\caption{Arrival rate estimates for varied true arrival rates.}
		\label{fig::varying_arrival_rates}
	\end{subfigure}
	\caption{The estimated arrival rate of the moving observer (blue) is compared against a stationary counter (red) and ground truth (black). The results are averaged over 100 simulated runs and the 90\% confidence intervals are shown.}
\end{figure}

To understand the effect of observation time on moving observer estimates, a simulation was run in which the number of times the vehicle traverses the link is varied.
Here the focus is on a single network link of length 100 m for which the vehicle only observes the link for one-third of the total simulation run time.
\cref{fig::varying_num_visits} shows how the estimates vary as a function of number of visits across the link.
For the 100 m link, the vehicle is able to correctly estimate after relatively few visits, but with low confidence in the estimates. 
With more visits to the link, the confidence interval decreases as is expected.

To understand the effect of true arrival rates on moving observer estimates, the same simulation is run over a range of true arrival rates with a nominal number of visits of 10.
\cref{fig::varying_arrival_rates} shows the estimated arrival rates from the moving observer method given different constant true arrival rates.
The results show that the arrival rate estimates from the moving observer method are equally valid over the range of arrival rates that are expected in the network graph according to \cref{fig::average_real_flow_rates}.

\section{Conclusion}\label{sec:conclusion}
 Having access to real time traffic arrival rate information is important for the management of an MOD system, but traditional methods of installing fixed sensors can be cost prohibitive.
This paper presented a method for estimating pedestrian arrival rates for links in an MOD network graph through the use of sensor equipped MOD vehicles. 
Pedestrian trajectory data is observed from MOD vehicles using a novel distributed fusion method which fuses LIDAR and camera data.
Benchmark testing revealed that the proposed distributed fusion method was able to outperform the standard maximum likelihood method, resulting in a 90\% hit rate for detecting pedestrian trajectories with 1.5 false positives per minute.
The observed pedestrian trajectories were used to determine the structure of the MOD network graph, and the framework for defining link arrival rates was presented.
A novel moving observer method was developed for estimating link arrival rates from pedestrian trajectory data.
Comparison between arrival rate estimates from the moving observer method and a stationary counter were made using real world experiments.
The moving observer method was shown to capture the time varying arrival rate trends that were also observed by the stationary counter.
A single moving observer was able generate arrival rate estimates for all links in the network graph, in both real world data and simulation, which otherwise could only have been possible through multiple, high-cost static link counters.

\section*{Acknowledgment}
Research supported by the Ford Motor Company through the Ford-MIT Alliance.

\balance

\bibliographystyle{IEEEtran}
\bibliography{IROS2016,biblio}

\addtolength{\textheight}{-0cm}
\end{document}